\documentclass{article}

\usepackage{arxiv}

\usepackage[utf8]{inputenc} % allow utf-8 input
\usepackage[T1]{fontenc}    % use 8-bit T1 fonts
\usepackage{hyperref}       % hyperlinks
\usepackage{url}            % simple URL typesetting
\usepackage{booktabs}       % professional-quality tables
\usepackage{amsfonts}       % blackboard math symbols
\usepackage{nicefrac}       % compact symbols for 1/2, etc.
\usepackage{microtype}      % microtypography
\usepackage{lipsum}

\usepackage{hyperref}
\usepackage{cleveref}

\usepackage{graphicx}
 \usepackage{algorithm}
\usepackage{algorithmic}

\usepackage{calrsfs}
\DeclareMathAlphabet{\pazocal}{OMS}{zplm}{m}{n}
\newcommand{\La}{\pazocal{L}}
\newcommand{\Ea}{\pazocal{E}}
\newcommand{\Fa}{\pazocal{F}}
\newcommand{\Aa}{\pazocal{A}}
\newcommand{\Ra}{\pazocal{R}}
\newcommand{\Sa}{\pazocal{S}}
\newcommand{\Ga}{\pazocal{G}}

\newcommand{\Pa}{\pazocal{P}}

\newcommand{\Ba}{\pazocal{B}}

\usepackage{amssymb,amsmath}

\title{Resolving Resource Incompatibilities in\\ Intelligent Agents}

\author{
  Mariela Morveli-Espinoza \\
  Graduate Program in Electrical and Computer Engineering (CPGEI),\\
 Federal University of Technology - Paran\'{a} (UTFPR),
 Curitiba - Brazil\\
  \texttt{morveli.espinoza@gmail.com} \\
   \And
   Ayslan Possebom \\
  Graduate Program in Electrical and Computer Engineering (CPGEI),\\
 Federal University of Technology - Paran\'{a} (UTFPR),
 Curitiba - Brazil\\
   \texttt{possebom@gmail.com} \\
   \And
   Cesar Augusto Tacla \\
   Graduate Program in Electrical and Computer Engineering (CPGEI),\\
 Federal University of Technology - Paran\'{a} (UTFPR),
 Curitiba - Brazil\\
 \texttt{tacla@utfpr.edu.br} \\
}

\newtheorem{defn}{Definition}

\newtheorem{exem}{Example}

\begin{document}

\maketitle

\begin{abstract}
An intelligent agent may in general pursue multiple procedural goals simultaneously, which may lead to arise some conflicts (incompatibilities) among them. In this paper, we focus on the incompatibilities that emerge due to resources limitations. Thus, the contribution of this article is twofold. On one hand, we give an algorithm for identifying resource incompatibilities from a set of pursued goals and, on the other hand, we propose two ways for selecting those goals that will continue to be pursued: (i) the first is based on abstract argumentation theory, and (ii) the second based on two algorithms developed by us. We illustrate our proposal using examples throughout the article.

\end{abstract}

% keywords can be removed

\section{Introduction}
An intelligent agent may in general pursue multiple procedural goals (hereafter, we will just call them goals) at the same time. In this situation, some conflicts among goals could arise, in the sense that it is not possible to pursue
them simultaneously. Thus, a rational agent should not simultaneously pursue a goal $g_i$ and a goal $g_j$ if $g_i$ prevents the achievement of $g_j$, in other words, if they are inconsistent \cite{winikoff2002declarative}. Since a rational agent requires that the goals pursued by him are consistent and do not conflict, he should be able to choose which of them will continue to be pursued, that is, he should be able to deal with such conflicts or incompatibilities. 

One type of incompatibility appears when the agent does not have the enough resources in order to achieve his pursued goals. For example, a cleaner robot agent has two pursued goals: (i) picking up a solid trash and (ii) helping another robot to pick up a big trash. For the first one, he needs 20 units of energy and for the second 40 units of energy, nevertheless he only has 50 units of such resource, therefore, a conflict between the two goals arises, and the agent has to choose which of them will continue pursuing.

Some related works have focused on conflicts definition and detection, as  Thangarajah \textit{et al.} \cite{thangarajah2002avoiding}, Van Riemsdijk \textit{et al.} \cite{van2009goals}, and Morveli \textit{et al.} \cite{morveli2017dealing}. Other ones worked on the strategy to choose the goals that the agent will continue to pursue, as Tinnemeier \textit{et al.} \cite{tinnemeier2007goal}, Pokahr \textit{et al.} \cite{pokahr2005goal}, and Wang \textit{et al.} \cite{wang2012runtime}. Finally, other works focus on the intention revision in order to detect conflict among a set of intentions and how to handle it, as Shapiro \textit{et al.} \cite{shapiro2012revising} and Zatelli \textit{et al.} \cite{zatelli2016conflicting}. 

One aim of this article is to show how to identify resource incompatibilities, and we also aim to deal with such possible incompatibilities. We start with a (possible incompatible) initial set of goals, on which we apply an algorithm to identify incompatible goals. The result of the incompatibility evaluation are, possibly, some sets of incompatible goals, which are assessed --based on the worth of each one for the agent-- in order to obtain a set of non incompatible goals.

In the next section, the mental states of the agent are presented. In Section \ref{recurso}, we define the resource incompatibility and show how this may be identified. Section \ref{evaluar} is devoted to the proposals for dealing with incompatible sets. Section \ref{related} discuss previous work in this area. Finally, last section summarizes our conclusions and points to possible future work.

\section{Agents}

In this work, an agent consists, basically, of a belief base $\Ba$, a set of goals $\Ga$, and a plan library $\Pa$. The belief base encodes the agent's knowledge about the world and $\Ga$ saves already pursued goals, which may be incompatible. Let $\La$ be the logical language used to represent such goals, beliefs and plans, and $\wedge$,$\vee$ and $\neg$ denote the logical connectives conjunction, disjunction and negation.

For the purposes of this paper we assume that an agent has a library of programmer-provided plans $\Pa$. Each plan consists of (i) an indicator that represents the goal for which it is relevant, (ii) a context condition which details the situations in which it is applicable, (iii) a plan body which specifies what the plan does, and (iv) a list of required resources, which stipulates the resources the plan needs in order to be executed. Thus, each plan has the following form: $pl_i=e : \psi \leftarrow P, [\Ra_{req}]$ encoding a plan-body program $P$ for handling an event-goal $e$ when the context condition $\psi$ is believed to hold, and with a resource requirements list $\Ra_{req}$, which is composed of a list of pairs $(res_i, n)$, where $n > 0$ represents the necessary amount of the resource $res_i$, for example $(energy, 200)$.

A resource summary $\Ra$ containing the information about the amount of resources the agent still has should also be defined. It is a list of pairs $(res_i, n)$, where $n > 0$ represents the total amount of the resource $res_i$ the agent still has, for example $\Ra=\{(energy, 200), (oil, 50), (fuel, 20)\}$. In both cases, for the list of required resources $\Ra_{req}$ and for the resource summary list $\Ra$, we assume that the resource sets are normalised so that each one appears exactly once.

%Finally, the agent needs to save a set of contrariness beliefs, which will be used in the identification of goals incompatibilities. $\Ca ontraries$ is the structure which saves such information, this is a list of pairs $(b_i, b_j)$, where $b_i, b_j \in \Ba$ and $b_i \neq b_j$. We write $b_i = -−b_j$ to say that $b_i = \neg b_j$ or $b_j = \neg b_i$, it means that $b_i$ and $- b_j$ are each other's negation. 

Finally, the agent is equipped with the following functions:

%\noindent - $\mathtt{GOAL}: \Pa lans \rightarrow \Ga$ is a function that returns the event-goal $e$ of a given plan $pl_i$. Notice that in this work $e$ is a goal.\\
%- $\mathtt{CONTEXT}: \Pa lans \rightarrow 2^{\La}$ is a functions returns the elements of the context of a given plan,\\
\noindent- $\mathtt{AVAILA\_RES}: \Ra \rightarrow \mathbb{R}$ is a function that returns the available quantity of a given resource the agent still has,\\
%- $\mathtt{RESOURCES}: \Ga \rightarrow 2^{\Ra}$ is a function that returns a list of necessary resources for achieving a given goal,\\
- $\mathtt{NEED\_RES}: \Ga \times \Ra \rightarrow \mathbb{R}$ is function that returns the amount of a given resource that a goal needs.

\section{Resource incompatibility}
\label{recurso}
Resource incompatibility is related to the necessary resources that goals need to be achieved. Thus, the quantity of resources of an agent can be enough for achieving a goal, nevertheless, when two or more goals need the same resources, it is possible that some conflicts arise.

%According to Thangarajah \textit{et al.} \cite{thangarajah2002avoiding}, the resources may be \textit{consumable}, i.e. they are no longer available after be used, or \textit{reusable}, i.e. following usage they are again available. For example energy is a consumable resource, whereas a communication channel can be considered a reusable resource.

%In order to reason about resource needs and how these affect the pursui of goals, we must develop a representation of resources which supports the desired reasoning. We assume a set of resource types T = ft1; : : : ; tng. For example Trover = fenergy;ComChg.

%We denote the consumable resources by $\Ra^c$ and the reusable ones by $\Ra^r$. Therefore, $\Ra = \Ra^c \cup \Ra^r$, such that $\Ra^c \cap \Ra^r = \emptyset$.

\subsection{Evaluating resources availability}
Before evaluating the resource incompatibility among goals, it must be checked whether there exist the enough amount of resources to achieve a given goal, the function in charge of this is the following:

\begin{itemize}

\item $\mathtt{EVAL\_RESOURCES}: \Ga \rightarrow 2^{\Ga}$ takes as input the set of goals of the agent and returns a maximal subset $\Ga' \subseteq \Ga$ of goals that are enabled to be analysed together. It means that the agent has enough resources for goals in $\Ga'$, but resources for goals in $\Ga - \Ga'$ are insufficient, whereby it is unnecessary to evaluate whether they have incompatibilities. Let us call goals in $\Ga'$ \textbf{enabled goals}.

\end{itemize}

\begin{exem} \label{ejm_res} Let $\Ra =\{(energy, 60), (time, 4)\}$ be the available quantity of resources of a given agent, and\break $\Ga=\{g_1, g_2, g_3\}$ be his pursued \hbox{goals. Let us} suppose that the necessary resource for executing a plan to achieve goal $g_1$ is $\{(energy, 70)\}$, for goal $g_2$ is $\{(energy, 35), (time, 2)\}$, and for goal $g_3$ is $\{(time, 3)\}$. Notice that there is not enough energy for executing the plan for goal $g_1$. However there are enough resources for executing either $g_2$ or $g_3$, therefore $\mathtt{EVAL\_RESOURCES}(\Ga)=\{g_2, g_3\}$.

\end{exem}

\subsection{Identifying resource incompatibility}
\label{identifica}
In order to determine resource incompatibilities among enabled goals that were returned by $\mathtt{EVAL\_RESOURCES}$, it must be analysed the requirements of goals in $\Ga'$, which may need either one or more resources in order to be achieved. There is resource incompatibility when two or more goals share the same need of a certain resource and the agent has no enough amount of such resource for all of them. Formally:

\begin{defn}\textbf{(Resource incompatibility)} Let $\Ga''=\{g_i, g_j, ..., g_k\}$ be a set of goals that need a same resource $res_m$ to be achieved, such that $ \Ga'' \subseteq \Ga' $. There is resource incompatibility among them when $(\sum_{x=i,j, ..., k} \mathtt{NEED\_RES}(g_x, res_m)) > \mathtt{AVAILA\_RES}(res_m)$.

%\begin{itemize}

%\item $\exists x \in \texttt{RESOURCES}(g_i)$ and $\exists y \in \texttt{RESOURCES}(g_j)$ such that $x=y$, and
%\item  

%\end{itemize}
\end{defn}
The following function is in charge of identifying the possible sets of incompatibilities goals: 

\begin{itemize}

\item $\mathtt{RESOURCE\_INCOM}(\Ga')$ takes as input the set of enabled goals $\Ga'$ and returns sets of incompatible goals. Let us call the returned set of sets of incompatible goals $\Sa_{incomp}$.

%When a the subset is singleton it means that  goal, otherwise it is returned as part of a set of incompatible goals and it is stored in the structure

\end{itemize}

Algorithm \ref{evalincomp} shows the performance of this function. The aim of the first loop (lines 2-4) is to generate a set for each internal resource of the agent, which is filled with enabled goals (lines 5-7). The idea is to put a given enabled goal in every set that corresponds to a resource the goal needs, thus, for example, if an enabled goal $g_2$ needs two resources $res_2$ and $res_6$, the sets generated for each of them must contain an element $g_2$. Finally, it is necessary that only sets with conflicting goals belong to $\Sa_{incomp}$ (lines 9-14). Hence, sets with one of the three possible characteristics are out of $\Sa_{incomp}$: (i) empty sets, (ii) singletons,  or (iii) sets where there is the necessary amount of resource to achieve all the goals that belong to it. %In relation to its time execution, it is linear --$O(n)$-- since it is proportional to the number of elements the set of enabled goals has. 

\begin{algorithm}
\label{evalincomp}
\begin{algorithmic}[1]
\REQUIRE A set of enabled goals $\Ga'$
\ENSURE A set of incompatible-goal sets $\Sa_{incomp}$
%\FORALL{$r_i \in \Ra esources$} Create an empty set $res_i$ \ENDFOR
\STATE{$\Sa_{incomp}= \emptyset$}
\FORALL {$res_i \in \Ra$}
    \STATE{Create an empty set $goals\_res_i$}
\ENDFOR
\STATE{$n=|\Ga'|$}
\FOR {$j=1$ to $n$} 
    \STATE{Put $g_j$ in each $goals\_res_i$ that corresponds to the resource it needs}
\ENDFOR
\STATE{$m=|\Ra|$}
\FOR {$i=1$ to $m$}
	\IF{$(\sum_{g_j \in goals\_res_i} \mathtt{NEED\_RES}(g_j, res_i))  > \mathtt{AVAILA\_RES}(res_i)$} 
		\STATE{$\Sa_{incomp}=\Sa_{incomp} \cup \{goals\_res_i\}$}
	\ENDIF
\ENDFOR
\RETURN {$\Sa_{incomp}$}
\end{algorithmic}
\caption{$\mathtt{RESOURCE\_INCOM}$: Identifying resource incompatibilities among goals}\label{evalincomp}
\end{algorithm}

Notice that in line 11 the algorithm does the sum for all the goals that need a certain resource, and when there is no enough resource for all the goals, they are considered incompatible ones. However, it could happen that there is enough resources for some of the goals that need such resource. For example, let us suppose that a given goal $g_i$ needs 20 units of energy, another goal $g_j$ 30 units, another goal $g_k$ 50 units, and the total energy of the agent is 50 units. It means that the three goals cannot be achieved, but the agent has enough energy for either achieving $g_i$ and $g_j$ or achieving only $g_k$. This situation is considered when the agent will choose the goals that he will continue to pursue and it is part of the next section.

\subsection{Simple incompatibility and complex incompatibility}
\label{tipos}

This subsection presents the types of incompatibility that can be detected over $\Sa_{incomp}$. These kinds of incompatibilities are related to the quantity of resource conflicts of a goal. Thus, when all goals in $\Sa_{incomp}$ have conflict only for one resource, we say that $\Sa_{incomp}$ has simple incompatibility, otherwise it has complex incompatibility.

Hereafter, (i) $\Ga_{incomp}$ is a set that contains all incompatible goals that belong to at least one of the subsets of $\Sa_{incomp}$, and (ii) $\Sa_j \in \Sa_{incomp}$ (for $1<j<|\Sa_{incomp}|$) is any set of incompatible goals.

%, for example, a goal $g_i$ has a conflict with another goal $g_j$ for resource time and with $g_k$ for resource money. W

\begin{defn} \textbf{(Simple incompatibility)} We say that $\Sa_{incomp}$ has simple incompatibility when $\forall g_i \in \Ga_{incomp}$ and $g_i \in \Sa_j$, such that $j$ has only one value, .
\end{defn}

\begin{defn} \textbf{(Complex incompatibility)} We say that $\Sa_{incomp}$ has complex incompatibility when $\exists g_i \in \Ga_{incomp}$ such that $g_i \in \Sa_j, ..., g_i \in \Sa_k$ where $\Sa_j, ..., \Sa_k \in \Sa_{incomp}$ and $j \neq ... \neq k$.  
\end{defn}

\begin{exem} Let $\Sa_{incomp}=\{\{g_8, g_4, g_5\}, \{g_9, g_6\}, \{g_9, g_7, g_8\}\}$.	
Notice that goals $g_4, g_5, g_6$ and $g_7$ have only one resource conflict, but goals $g_8$ and $g_9$ have more than one. Therefore, $\Sa_{incomp}$ has complex incompatibility.

\end{exem}

\section{Evaluating incompatible goals}
\label{evaluar}
In this section, we present the process for solving incompatibilities, and with solving we mean to choose which goals the agent can continue to pursue. With this aim, we propose two forms for solving such incompatibilities, our first proposal is based on abstract argumentation and the second one is based on two algorithms developed by us. The result of both proposals is a set of goals that can continue to be pursued. Let us call this set $\Ga_{cons}$ (we use this notation because these are consistent goals).

In both proposals, the assessment of the goals depends on how valuable a goal is for the agent. According to Castelfranchi and Paglieri \cite{castelfranchi2007role}, when there exist incompatibilities among goals, the agent should determine how valuable is each of them for him. Thus, the most valuable ones can continue to be pursued. In this work, we assume an agent is equipped with a function that returns goals' values:

\begin{itemize}

\item $\mathtt{WORTH}:\Ga \rightarrow [0,1]$ (values in $[0,1]$ belong to $\mathbb{R}$) is the function  that returns a value for every goal in $\Ga$, where 0 is the minimum value a goal may have and 1 the maximum one. This function have to be applied to every goal in $\Ga$, therefore, each goal has one value in order to be evaluated.

\end{itemize}

Finally, let us point out that there is a total order on the goals' values.

\subsection{Proposal based on abstract argumentation theory}
We present a framework for solving incompatibilities using abstract argumentation. More precisely given a set of incompatible goals, we are going to use abstract argumentation theory for deciding which of them will continue to be pursued. Unlike the next proposal (see next subsection), in this one, it is not taken into account the differentiation of types of incompatibility.

In abstract argumentation theory of Dung \cite{dung1995acceptability}, an argumentation framework ($\Aa\Fa$) is composed by a set of arguments and a binary attack relation defined over the set.  An $\Aa\Fa$ can also be represented by a directed graph where the nodes represent the arguments and the edges the attacks. Generalizing, an argumentation framework is a set of elements with conflicting relations among them. Comparing with the problem of incompatibility among goals, we can notice that it  has also the same characteristics, the set of elements would be the goals and the incompatibility the conflicting relation.
Based on this, we propose to model the incompatibility among goals using this theory.

\begin{defn} \textbf{(Incompatible goals framework)} An argumentation-like framework for dealing with incompatibility among goals is a triple $\Ga\Fa =\langle \Ga_{incomp}, \Ra_{incomp} ,\mathtt{WORTH} \rangle$, where:

\begin{itemize}
\item $\Ga_{incomp}$ is a set of incompatible goals,
\item $\Ra_{incomp} \subseteq  \Ga_{incomp} \times \Ga_{incomp}$ is a symmetric binary relation of incompatibility.  A goal $g_i \in \Ga_{incomp}$ is incompatible with another goal $g_j \in \Ga_{incomp}$ if $(g_i, g_j) \in  \Ra_{incomp}$. Besides, $(g_j, g_i) \in  \Ra_{incomp}$ since incompatibility is a symmetric relation. These relations are obtained from the set $\Sa_{incomp}$,
\item $\mathtt{WORTH}$ is a function that returns the value of a given goal.
\end{itemize}
 
\end{defn}

\begin{figure}[!htb]
	\centering
	\includegraphics[width=0.40\textwidth]{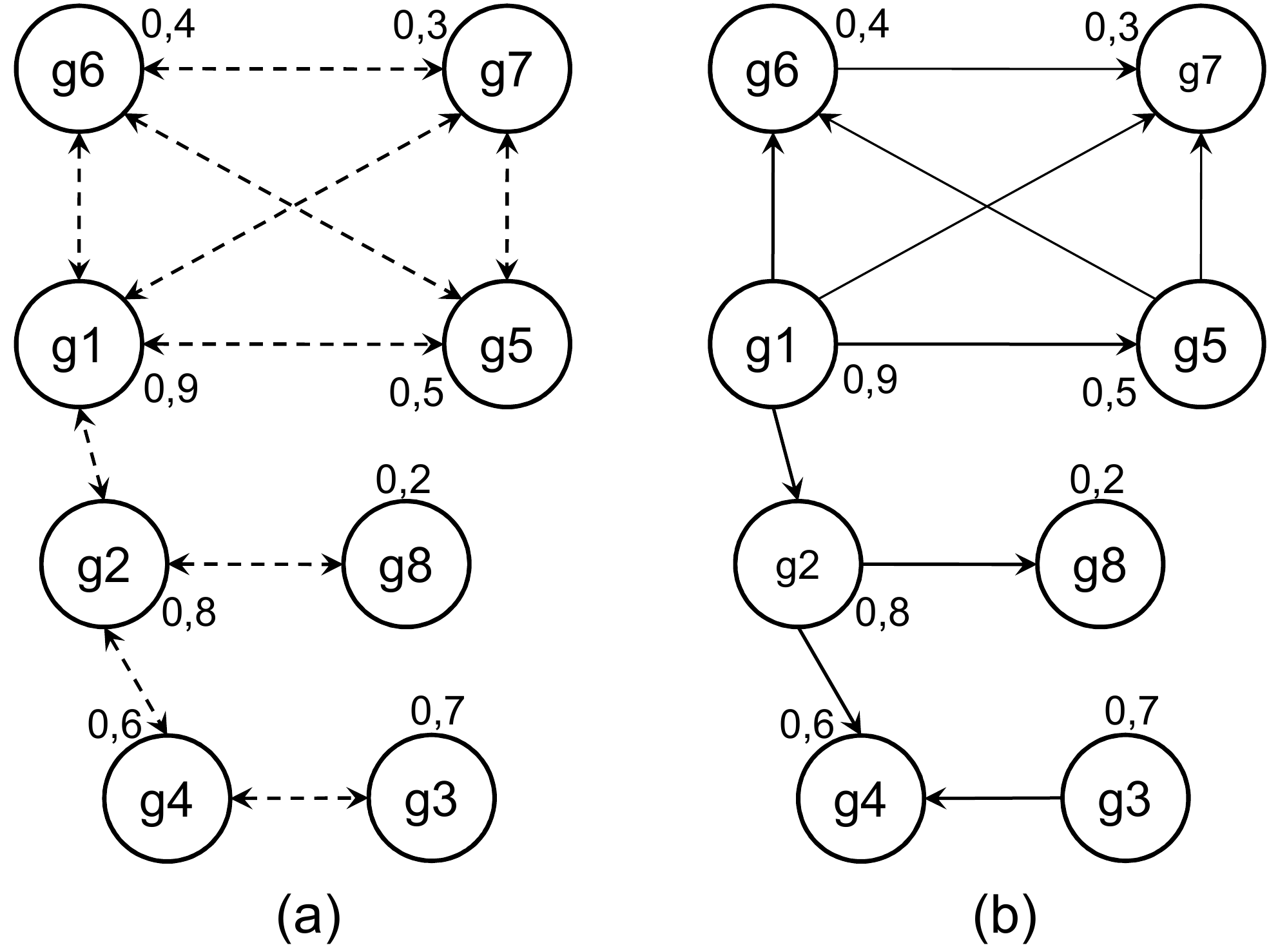} % <- formatos PNG, JPG e PDF
	\caption{(a) Graph representation of the incompatibilities among goals. (b) Graph representation considering the defeat relation.}
	\label{grafo}
\end{figure}

\begin{exem} \label{ejm1} Let $\Sa_{incomp}= \{\{g_1, g_2\}, \{g_2, g_4\}, \{g_3, g_4\}, \{g_1, g_5, g_6, g_7\}, \{g_2, g_8\}\}$ be a set of sets of incompatible goals, hence $\Ga_{incomp}=\{g_1, g_2, g_3, g_4, g_5, g_6, g_7, g_8\}$. This can be translated into a  $\Ga\Fa$ represented by the directed graph in Figure \ref{grafo}(a), where nodes represent the goals and edges the incompatible relation between goals. Numbers next to the nodes represent the value returned by function $\mathtt{WORTH}$ for each goal. All edges have double direction due to the symmetric nature of relation $\Ra_{incomp}$.
\end{exem}

Since the attack relation is symmetric, we will use the value of each goal to break that symmetry in order to decide which attacks succeed as defeats.

\begin{defn} \textbf{(Defeat)} Given an attack relation $(g_i, g_j)$, goal $g_i$ defeats goal $g_j$ only when $\mathtt{WORTH}(g_i) > \mathtt{WORTH}(g_j)$, for $j \neq i$.
\end{defn}

\begin{exem}Continuing Example \ref{ejm1}, the graph presented in Figure \ref{grafo}(b) is the resultant taking into account the values of goals. Notice that all attacks only have one direction.

So far, we have presented the representation of goals incompatibility, the next step is to identify which of these goals will continue to be pursued, to which end we will use the concept of semantics. 

In abstract argumentation theory several semantics of \textbf{acceptability} were defined \cite{dung1995acceptability}. These produce none, one or several acceptable sets of arguments, called extensions, which contain a set of consistent arguments. In our case, semantics will be used to produce a set of non-conflicting goals. But besides that we aim to obtain a set of the most valuable non-conflicting goals, which is guaranteed since a goal only defeats another one if it has a greater value.

\end{exem}

Following some definitions related to the semantics, adapted to our work:

\begin{defn} \textbf{(Conflict-freeness)} Let $\Ga\Fa=\langle \Ga_{incomp}, \Ra_{incomp} ,\mathtt{WORTH} \rangle$ be an incompatible goals framework and $\Ea \subseteq \Ga_{incomp}$. $\Ea$ is conflict-free iff there exist no $g_i, g_j \in \Ea$ such that $g_i$ defeats $g_j$.

\end{defn}

\begin{defn}\textbf{(Semantics)} Let $\Ea \subseteq \Ga_{incomp}$:

\begin{itemize}

\item $\Ea$ \textbf{defends} a goal $g_i$ iff for each goal $g_j \in \Ga_{incomp}$, if $g_j$ defeats $g_i$, then there exist a goal $g_k \in \Ea$ such that $g_k$ defeats $g_j$.
\item $\Ea$ is \textbf{admissible} iff it is conflict-free and defends all its elements.
\item A conflict-free $\Ea$ is a \textbf{complete extension} iff we have $\Ea = \{g_i | \Ea$ defends $g_i\}$.
\item $\Ea$ is a \textbf{preferred extension} iff it is a maximal (for set inclusion) complete extension.
\item $\Ea$ is a \textbf{grounded extension} iff it is the smallest (for set inclusion) complete extension.
\end{itemize}

\end{defn}

The role of the semantics is to define which goals will be considered acceptable, in our case, which goals  are consistent. Preferred extensions are specially useful when two goals have the same value and the grounded one when the values are different. The advantage of the grounded extension is that it always exist and it is unique for  every $\Ga\Fa$. In the case that the grounded extension is an empty set, we may use the
preferred extension. 

\begin{defn}\textbf{(Acceptable goal)} A goal $g_i$ is considered acceptable iff $g_i \in \Ea$ such that $\Ea$ is a preferred or a grounded extension. All acceptable goals belong to the set of consistent goals $\Ga_{cons}$.

\end{defn}

\begin{exem} In order to calculate the semantics for Example \ref{ejm1} we will use ConArg \cite{bistarelli2011modeling}, a computational tool for modeling and solving argumentation frameworks. 

Figure \ref{grafpref}(a) shows the resultant grounded extension $\Ea=\{g_1, g_3, g_8\}$, which means that such goals are considered acceptable or consistent ones, and will continue to be pursued. 

Notice that although $g_2$ is more valuable than $g_8$, it is no considered acceptable, this is due to $g_1$, whose value is the greatest one, and since $g_1$ defeats $g_2$, it cannot be considered part of the grounded extension. 

%\begin{figure}[!htb]
%	\centering
%	\includegraphics[width=0.35\textwidth]{grafo-gro.pdf} % <- formatos PNG, JPG e PDF
%	\caption{Grey filled nodes represent the grounded extension and hence the acceptable goals. Thus $%\Ga_{cons}=\{g_1, g_3, g_8\}$}.
%	\label{grafgro}
%\end{figure}

\end{exem}

\begin{exem} Above we have claimed that preferred extension are specially useful when goals have the same value. Let us suppose that all goals of previous examples have the same value. In such case the resultant grounded extension is $\Ea=\emptyset$. However using preferred extensions the agent obtains two subsets of acceptable goals: $\{g_1, g_8, g_3\}$ and $\{g_1, g_8, g_4\}$, from which he  can choose one and therefore obtain a set of acceptable goals. Figure \ref{grafpref}(b) shows both preferred extensions.

\begin{figure}[!htb]
	\centering
	\includegraphics[width=0.5\textwidth]{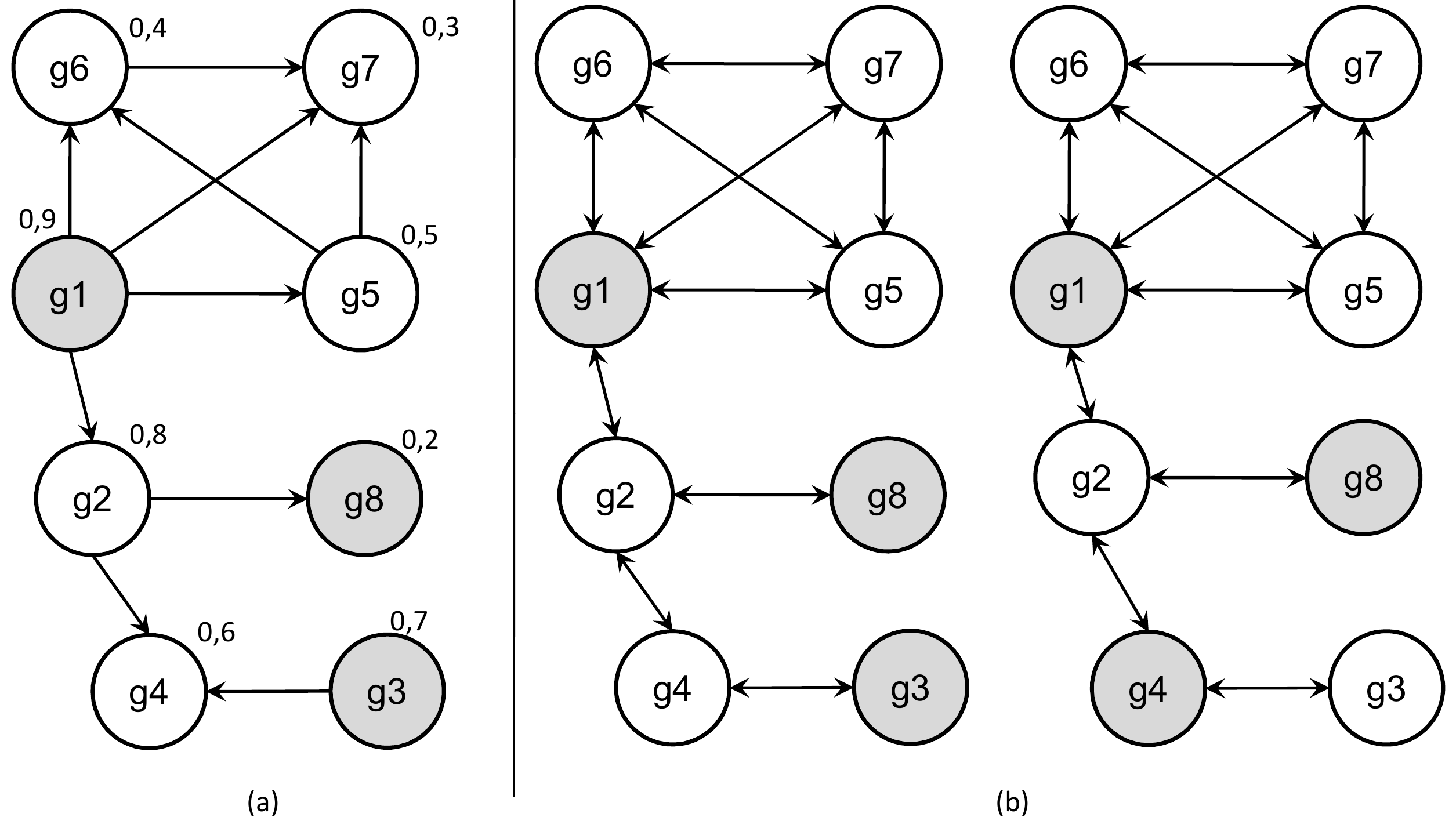} % <- formatos PNG, JPG e PDF
	\caption{(a) Grey filled nodes represent the grounded extension and hence the acceptable goals. (b) Grey filled nodes represent the goals that belong to each preferred extension.}
	\label{grafpref}
\end{figure}

\end{exem}

One drawback of this proposal is that it does not deal with the situation described at the end of Subsection \ref{identifica}. Let us recall that this situation happens when the agent does not have enough quantity of a resource for achieving every goal that needs it, but this quantity can let him to achieve some of the goals. For dealing with it, we propose two algorithms that are described in the following subsection. 

\subsection{Algorithm-based proposal}

In this sub-section two algorithms will be presented, one for dealing with simple incompatibility and the other one for dealing with complex incompatibility.

Depending on the type of incompatibility of $\Sa_{incomp}$, two cases can be distinguished.% Although three possible combinations where presented in Section \ref{tipos}, these can be reduced to just two cases.

%\begin{itemize}

\textbf{Case 1:} When $\Sa_{incomp}$ has simple incompatibility, the set of goals that will continue to be pursued, i.e. the consistent goals set is made up of the most valuable goals of each subset of $\Sa_{incomp}$.

\textbf{Case 2:} When $\Sa_{incomp}$ has complex incompatibility, one of the three possible cases may occur. 

Before enumerate the cases, let $g_i$ be a goal that belongs to more than one subset of $\Sa_{incomp}$, let $\Ga_{>} \subseteq \Ga_{incomp}/g_i$ represent the set of goals that have incompatibilities with $g_i$ and are more valuable than $g_i$, and  let $\Ga_{<}\subseteq \Ga_{incomp}/g_i$ represent the set of goals that have incompatibilities with $g_i$ and are less valuable than it. Now, let's see the three  possible situations:

%\begin{enumerate}
1) When $\Ga_{<}= \emptyset$, it means that there is no goal more valuable than $g_i$. In this case, $g_i \in \Ga_{cons}$, i.e $g_i$ makes part of the set of goals that will continue to be pursued.

2) When $\Ga_{>}= \emptyset$, it means that all goals that have incompatibilities with $g_i$ are more valuable than it. In this case, $g_i \notin \Ga_{cons}$.

3) When there is unless one $g_j \in \Ga_{<}$ and one $g_k \in \Ga_{>}$, then $g_i$ may belong to $\Ga_{cons}$ or not. For example, let us suppose that $g_i$ has incompatibility with $g_j$ about resource $res_1$ and with $g_k$ about resource $res_2$, let $\mathtt{WORTH} (g_j) > \mathtt{WORTH} (g_i) > \mathtt{WORTH}(g_k)$. If after deducting the resource(s) for $g_j$, there is still enough resource(s) for $g_i$, then it can be part of $\Ga_{cons}$. Otherwise, it would be useless to obtain just part of the resources $g_i$ needs to be achieved. This also means that it is possible that some $g_k \in \Ga_{>}$ be part of $\Ga_{cons}$ even when it is less valuable. 
%\end{enumerate}

%\end{itemize}
In order to deal with this cases, the following functions are defined:

\begin{itemize}
\item $\mathtt{EVAL\_SIMPLE}: \Sa_{incomp} \rightarrow \Ga_{cons}$
\item $\mathtt{EVAL\_COMPLEX}: \Sa_{incomp} \rightarrow \Ga_{cons}$
\end{itemize}

Both functions take as input $\Sa_{incomp}$ and return a set of consistent goals. The difference is that the first one is used to deal with Case 1, i.e. when $\Sa_{incomp}$ has simple incompatibility, and the second one for dealing with Case 2. Below, we present the algorithm for each one.

\begin{algorithm}
\begin{algorithmic}[1]
\REQUIRE $\Sa_{incomp}$ with simple incompatibility
\ENSURE A set of consistent goals $\Ga_{cons}$
%\FORALL{$r_i \in \Ra esources$} Create an empty set $res_i$ \ENDFOR
%\STATE{$most\_pref=g_0$}
\STATE{$\Ga_{cons}=\{\}$} 
\STATE{$\mathtt{WORTH}(most\_val)=-1$} 
\STATE{$n=|\Sa_{incomp}|$}
\FOR{$i=1$ to $n$}
	\STATE{$m=|\Sa_{incomp}[i]|$}
	\FOR{$j=1$ to $m$} 
		\IF{$\mathtt{WORTH}(g_j)>\mathtt{WORTH}(most\_val)$}
			\STATE{$most\_val=g_j$}
		\ELSIF {$\mathtt{WORTH}(g_j)==\mathtt{WORTH}(most\_val)$}
			\STATE{$most\_val=\mathtt{RANDOM}(g_j, most\_val)$}
		\ENDIF

	\ENDFOR	
	\STATE{$\Ga_{cons}=\Ga_{cons} \cup most\_val$}	
	\STATE{$\mathtt{WORTH}(most\_val)=-1$} 
\ENDFOR

\RETURN {$\Ga_{cons}$}
\end{algorithmic}
\caption{$\mathtt{EVAL\_SIMPLE}$: Evaluating simple incompatibility}\label{evalsimple}
\end{algorithm}

Algorithm \ref{evalsimple} shows the behaviour of function $\mathtt{EVAL\_SIMPLE}$, and Algorithm \ref{evalvalue} the behaviour of $\mathtt{EVAL\_COMPLEX}$.

The idea of Algorithm \ref{evalsimple} is to evaluate each subset of $\Sa_{incomp}$ and the most valuable goal this is added to the set $\Ga_{cons}$. In the case of Algorithm \ref{evalvalue}, it evaluates each goal of $\Ga_{incomp}$ (function $\mathtt{TAKE\_ELEM}$ returns an element from $\Ga_{incomp}$). From line 4 to line 16, the algorithm evaluates each subset of $\Sa_{incomp}$, if the evaluated goal is the most valuable of each subset and there is enough resources for achieving it in every subset, then it is added to $\Ga_{cons}$. After its evaluation, the goal is removed from $\Ga_{incomp}$ and from every subset of $\Sa_{incomp}$ it belongs (lines 20 and 21). Finally, function $\mathtt{CLEAN}$ removes from $\Sa_{incomp}$ possible empty subsets.

\begin{algorithm}
%\label{evalvalue}
\begin{algorithmic}[1]
\REQUIRE The set $\Sa_{incomp}$
\ENSURE A set of consistent goals $\Ga_{cons}$

  \REPEAT
    \STATE {$g=\mathtt{TAKE\_ELEM}(\Ga_{incomp})$} 
    \STATE{flag=0}
    \FOR{$i=1$ to $|\Sa_{incomp}|$}
    	\IF{$g \in \Sa_{incomp}[i]$}
    		\IF{($\mathtt{MOST\_VAL}(g, \Sa_{incomp}[i])$)}
	    		\IF{($\mathtt{ENOUGH\_RES}(g, res_i)$)}
					\STATE{flag=1}
				\ELSIF{}
					\STATE{flag=0}		
    			\ENDIF
    		\ELSIF{}
    			\STATE{flag=0}		
			\ENDIF
    	\ENDIF
    \ENDFOR
    \IF{(flag==1)}
    	\STATE{$\Ga_{cons}=\Ga_{cons} \cup g$}
    \ENDIF
    \STATE{$\Ga_{incomp}=\Ga_{incomp}-g$}
    \STATE{$\Sa_{incomp}= \mathtt{DELETE}(\Sa_{incomp},g)$}
	\STATE{$\mathtt{CLEAN}(\Sa_{incomp})$}
  \UNTIL{$\Ga_{incomp}=\emptyset$} % <--- use \doWhile for the "while" at the end

\RETURN {$\Ga_{cons}$}
\end{algorithmic}
\caption{$\mathtt{EVAL\_COMPLEX}$: Evaluating sets of incompatible goals}\label{evalvalue}
\end{algorithm}

\begin{exem} Let $\Sa_{incomp}=\{\{g_1, g_2, g_8\}, \{g_2, g_6\}, \{g_1, g_4\}, \{g_7, g_9\}\}$, hence $\Ga_{incomp}=\{g_1, g_2, g_4, g_6, g_7, g_8,\break g_9\}$. Let $\mathtt{WORTH}(g_1)>\mathtt{WORTH}(g_2)>\mathtt{WORTH}(g_4)>\mathtt{WORTH}(g_6)>\mathtt{WORTH}(g_7)>\mathtt{WORTH}(g_8)>\mathtt{WORTH}(g_9)$. 

\begin{table}[htb]
\label{ejemplo}
\caption{Necessary resources for each goal. Last column shows the available resource of the agent.}
\begin{center}
\begin{tabular}{|c|c|c|c|c|c|c|c|c|c|}
 \hline 
 RES. & $g_1$ & $g_2$ & $g_4$ & $g_6$ & $g_7$ & $g_8$ & $g_9$ & AVA.\\ 
 \hline 
 $res_A$ & 30 & 20 & 0 & 0 & 0 & 50 &0&50\\ 
 \hline 
 $res_B$ & 0 & 10 & 0 & 7 & 0 & 0 &0&10\\ 
 \hline 
 $res_C$ & 5 & 0 & 7 & 0 & 0 & 0 &0&10\\ 
 \hline 
 $res_D$ & 0 & 0 & 0 & 0 & 15 & 0 &20&25\\ 
 \hline 

 \end{tabular}  

\end{center}
\end{table}
Notice that $res_A$ is enough either for goals $g_1$ and $g_2$ or for $g_8$. The final allocation of the resource depends on the worth of goals that need it. After applying Algorithm \ref{evalvalue}, the set of consistent goals is $\Ga_{cons}=\{g_1, g_2, g_7\}$.
\end{exem}

%Let's see the importance of incompatible goals returned by function $EvalIncomp$ in the Example \ref{ejm_incom}:
%$Importance(cook\_lasagna(workers\_day))= 0.8$
%$Importance(prepare\_lemon\_pie(workers\_day))= 0.6$\\

%Taking into account only the importance, the result of function $EvalValue$ is: 

%$EvalValue(cook\_lasagna(workers\_day), prepare\_lemon\_pie(Workers\_day))\linebreak = cook\_lasagna(workers\_day)$. 

%Hence, a value belief is created for it and $\Ba^* = \Ba^* \cup \linebreak \{most\_valuable('cook\_lasagna(workers\_day)')\}$.

\section{Related works}
\label{related}

Thangarajah \textit{et al.} \cite{thangarajah2002avoiding} focus on the detection of resource conflict, they characterise different types of resources and present mechanisms that allow agents to be aware of the possible and necessary resource requirements. 

Van Riemsdijk \textit{et al.} \cite{van2009goals} formalize two ways of representing conflicting goals, using propositional and default logic. They argued that even logically consistent goals can be conflicting, for instance, when multiple goals/plans are chosen to fulfill the same (super)goal.

Other approaches worked on the deliberation strategy. Tinnemeier \textit{et al.} \cite{tinnemeier2007goal} propose a mechanism to process incompatible goals by refraining the agent from adopting plans for goals that hinder the achievement of goals the agent is currently pursuing. They also define three types of goal selection strategies depending on the priority of the goals. Pokahr \textit{et al.} \cite{pokahr2005goal} propose a goal deliberation strategy called Easy
Deliberation, which allows agent programmers to specify the relationships
between goals and enables an agent to deliberate about its goals by activating and deactivating certain goals. Their strategy enforces that only conflict free goals are pursued and considers the order of goal importance. Wang \textit{et al.} \cite{wang2012runtime} proposes a runtime goal conflict resolution model for agent systems, which consists of a goal state transition structure and a goal deliberation mechanism based on a  extended event calculus.

Finally, Khan and Lesp{\'e}rance \cite{khan2010logical} present a framework where an agent can have multiple goals at different priority levels, possibly inconsistent with each other. The information about the goals priority is used to decide which of the them should no longer be actively pursued in case they
become mutually inconsistent. In their approach lower priority goals are not dropped permanently
when they are inconsistent with other goals but they become inactive ones. In this way, when the world changes, the agent recomputes her adopted goals and some inactive goals may become active again.
They argue that this ensures that an agent maximizes her utility.

\section{Conclusions and future work}
The first part of this work presents the formalization and identification of resource incompatibility among goals. Although in \cite{thangarajah2002avoiding} the types of resources and how to avoid conflicts has been addressed, it has not be showed when such conflicts arises; and in \cite{van2009goals} they formalize how to represent conflicting goals, however they do not show how to identify when two goals have conflicts either.

As in \cite{tinnemeier2007goal, pokahr2005goal} and \cite{khan2010logical} we have considered the worth of a goal to deal with emerging conflicts. Nevertheless, we have not studied the criteria that may be taken into account (e.g. goal importance or goal urgency) inside function $\mathtt{WORTH}$, whereby in order to obtain a more accurate value that really represents the preferences and needs of the agent, we plan to study it more deeply.

We have also identified two types of incompatibility, which were defined considering the resultant sets of incompatible goals. From these types, we observed that the strategy for selecting the goals that will continue to be pursued (or consistent goals) may be different. Thus, we developed an algorithm for dealing with simple incompatibility and another one for complex incompatibility. 

Besides, we noticed that the problem of selecting a set of goals from a larger set of incompatible ones can be compared to the problem of calculating an extension in abstract argumentation. Therefore we have adapted concepts of abstract argumentation to our problem. Unlike the case of algorithm-based proposal, it is not necessary to consider the types of incompatibility. An example is provided to demonstrate how this theory can be applied.

We have studied this problem in the goals level, however, it would be more interesting to study it at the plans level. This is in fact, an interesting direction that we will follow.

Finally, unlike \cite{tinnemeier2007goal} and \cite{khan2010logical}, this work does not deal with already adopted goals, it will be for sure another interesting direction of our future research.
%\end{document}

% conference papers do not normally have an appendix

% use section* for acknowledgment
\section*{Acknowledgment}
This work is fully founded by CAPES (Coordena\c{c}\~{a}o de Aperfei\c{c}oamento de Pessoal de N\'{i}vel Superior).

\bibliographystyle{unsrt}  
\bibliography{IEEEtran}  %%% Remove comment to use the external .bib file (using bibtex).
%%% and comment out the ``thebibliography'' section.

%%% Comment out this section when you \bibliography{references} is enabled.

\end{document}